\documentclass[preprint,12pt]{elsarticle}
\usepackage{times}
\usepackage{epsfig}
\usepackage{graphicx}
\usepackage{amsmath}
\usepackage{amssymb}
\usepackage{multirow}
\usepackage{epstopdf}
\usepackage{caption}
\usepackage{subfigure}
\usepackage{tabularx}
\usepackage{algorithm,algpseudocode}
\usepackage{float}
\usepackage{amssymb}
\usepackage{gensymb}
\usepackage{booktabs}

\algdef{SE}[DOWHILE]{Do}{doWhile}{\algorithmicdo}[1]{\algorithmicwhile\ #1}%

\usepackage{enumitem}

\begin{document}

\begin{frontmatter}

\title{Can We Automate Diagrammatic Reasoning?}

\author[ska]{Sk. Arif Ahmed}
\ead{arif.1984.in@ieee.org} \fntext[ska]{Haldia Institute of
Technology, Haldia, India}

\author[dpd]{Debi Prosad Dogra}
\ead{dpdogra@iitbbs.ac.in} \fntext[dpd]{Indian Institute of
Technology, Bhubaneswar, India}

\author[skar]{Samarjit Kar}
\ead{samarjit.kar@maths.nitdgp.ac.in} \fntext[skar]{National
Institute of Technology, Durgapur, India}

\author[proy]{Partha Pratim Roy}
\ead{proy.fcs@iitr.ac.in}
 \fntext[proy]{Indian Institute of Technology, Roorkee, India}

\author[dkp]{Dilip K. Prasad}
\ead{dilipprasad@gmail.com }
 \fntext[dkp]{UiT-The Arctic Uninversity of Norway, Tromsø,Norway}

\begin{abstract}
Learning to solve diagrammatic reasoning (DR) can be a
challenging but interesting problem to the computer vision research community. It is believed that next generation pattern recognition applications should be able to simulate human brain to understand and analyze reasoning of images. However, due to the lack of benchmarks of diagrammatic reasoning, the present research primarily focuses on visual reasoning that can be applied to real-world objects. In this paper, we present a diagrammatic reasoning dataset that provides a large variety of DR problems. In addition, we also propose a Knowledge-based Long Short Term Memory (KLSTM) to solve diagrammatic reasoning problems. Our proposed analysis is arguably the first work in this research area. Several state-of-the-art learning frameworks have been used to compare with the proposed KLSTM framework in the present context. Preliminary results indicate that the domain is highly related to computer vision and pattern recognition research with several challenging avenues.

\end{abstract}


\end{frontmatter}

\section{Introduction}
\begin{figure}[!htb]
\centering
\includegraphics[scale=0.6]{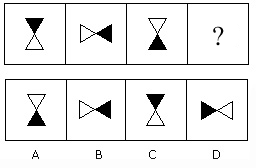}
\caption{A typical example of a diagrammatic reasoning problem. The first row presents the first three objects of a sequence of four objects in a particular order. The second row presents the multiple choices typically shown to an examinee. Option D is the right answer for the above problem.}\label{fig:teas}
\end{figure}

Diagrammatic reasoning involves visual representations of objects or diagrams. It involves understanding concepts and ideas from images consisting of patterns. Solving such diagrammatic reasoning problems using computer vision and artificial intelligence can help us to understand complex patterns of objects in images. Typically, a test in diagrammatic reasoning consists of questions that requires analyzing a sequence of shapes or patterns. This is also known as abstract or inductive reasoning test[]. The task is to identify the rules that can be applied to a sequence and then use them to pick an appropriate answer. The questions are usually of multiple choices. These questions generally consist of a series of pictures, each of which is different or oriented. The task is to choose another picture from a number of options to complete the series. For example, Figure~\ref{fig:teas} shows a typical diagrammatic reasoning problem, where the first row represents the question and the second row contains the four options out of which only one is correct.
\subsection{Related Work}
\begin{figure}[!htb]
\centering
\includegraphics[scale=0.4]{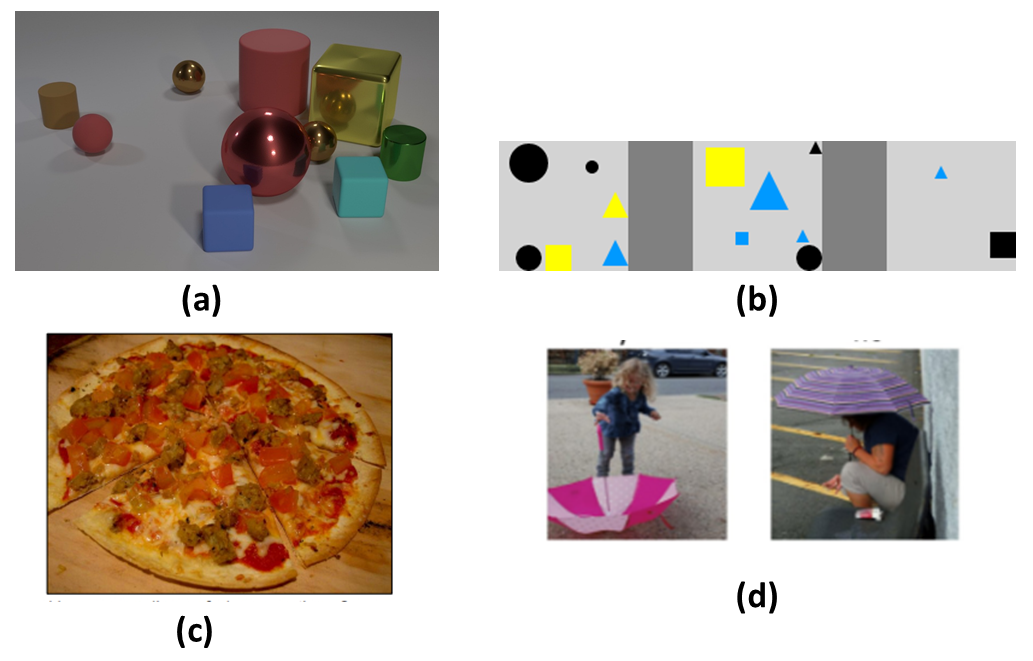}

\caption{\textbf{Visual reasoning dataset.} (a) Question: How many objects are either small cylinders or red things? (b) Question: There is exactly one big yellow square not touching any edge (True/False) (c) Q: How many slices are there in the pizza? (d)  Q: Is the umbrella upside down?}
\label{fig:related1}
\end{figure}

Solving reasoning problems using artificial intelligence (AI) is a challenging task. For example, solving mathematical word problems~\cite{kushman2014learning} using natural language processing (NLP) is well-known in artificial intelligence. Solutions to them have enhanced the strategies of supervised learning by introducing newer rules. However, similar tasks in visual reasoning have not received focused attention of the computer vision research community. Two similar domains  that have attracted computer vision and pattern recognition researchers are visual question answering~\cite{antol2015vqa,yang2016stacked,noh2016image} and visual reasoning~\cite{johnson2017inferring,hu2017learning,hu2017modeling}. Figures~\ref{fig:related1}(a) depicts Compositional Language and Elementary Visual Reasoning (CLVR)~\cite{johnson2017clevr} dataset that has been used to build artificial intelligence systems to reason and answer questions about visual data. Figures~\ref{fig:related1}(b) depicts Cornell Natural Language Visual Reasoning (NLVR) synthetic dataset to solve the task of determining whether a comment is true or false about an image, Figures~\ref{fig:related1}(c) represents Visual Question Answering (VQA) dataset. VQA is a new dataset containing open-ended questions about the images~\cite{antol2015vqa}. Figures~\ref{fig:related1}(d) represents reasoning of image pairs~\cite{balanced_vqa_v2}.

\subsection{Motivation and Contributions}
Diagrammatic reasoning can also be presented as a visual sequence prediction problem~\cite{gavornik2014learned}. In computer vision, similar approaches are used to predict future video frames~\cite{lu2017flexible,liu2017video,vukotic2017one}.
Following a similar line of thinking like visual reasoning in computer vision, we have introduced this new domain of research, namely solving diagrammatic reasoning using machine learning guided computer vision process. In this context, we have made the following contributions:
\begin{itemize}
  \item Solving diagrammatic reasoning problems with the help of computer vision and pattern recognition techniques.
  \item Introducing  a rich diagrammatic reasoning dataset that can be used by the computer vision research community for solving similar problems through pattern recognition and machine learning.
  \item We also introduce a new learning framework referred to as Knowledge-based Long Short Term Memory (KLSTM) to solve diagrammatic reasoning problems.
\end{itemize}
Rest of the paper is organized as follows. In Section 2, we present the Datasets and Benckmarks. Section 3 presents the proposed DR solving method. Experiment results are presented in Section 4. Conclusion and future work are presented in Section 5.
\section{Datasets and Benchmarks}
\begin{figure}[h]
\centering
\includegraphics[scale=0.6]{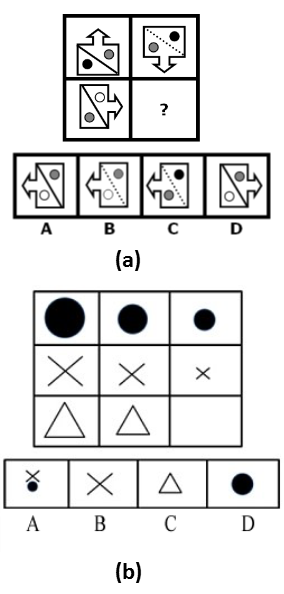}
\caption{Examples of DR problems that are difficult in nature and may not be possible to solve using typical machine learning frameworks. (a) Example of a $2\times2$ matrix reasoning problem. (b) Example of a $3\times3$ graphical reasoning problems.}\label{fig:dataset2}
\end{figure}

The ultimate goal of visual reasoning is to learn image understanding and interpretations. Due to the unavailability of datasest and benchmarks, research in this domain is still in its infancy. There are a large variety of DR  problems. For examples, Figure~\ref{fig:dataset2}(a) represents a $2\times2$ and Figure~\ref{fig:dataset2}(b) represents a $3\times3$ DR problem. The examples seem complex and we have found these examples are hard to solve through common machine learning frameworks. Such problems are left for future research. In this paper, we have considered a $4\times1$ diagrammatic problem, such as shown in Figure~\ref{fig:teas}.

\begin{figure*}[!htb]
\centering
\includegraphics[scale=0.5]{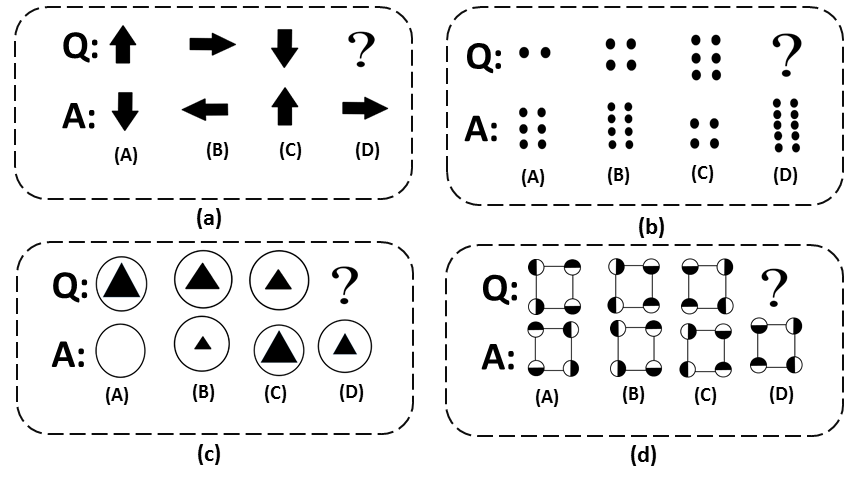}
\caption{\textbf{Examples of four types of typical DR problems that are present in our dataset.} (a) Example of a rotation problem (RT), where a pattern is rotated as compared to the first image with relative rotations that may be mentioned in a DR question as $\{0\degree,90\degree,180\degree,?\}$. The prediction should be $270\degree$ and the correct answer is option B. (b) It is a typical problem of number series prediction (CT). The question consists of  a set of filled circles. Here, the number of circles varies as 2, 4, 6, ?. Our task is to predict the picture with  8 filled-circles. The correct answer is option B. (c) Third one is an example of typical shape and scaling problem (SS). The pattern can be interpreted as \{$<$Cicle, Large Triangle$>$,$<$Circle, Big Triangle$>$, $<$Circle, Small Triangle$>$,$<$?$>$\}. Our task is to predict $<$Circle, Tiny Triangle$>$ which is option B. (d) The fourth one is a typical pattern understanding problem. We have categorized such problems into Other Type (OT). Our task is to predict the $4^{th}$ pattern. The correct answer is option A.}
\label{fig:dataset1}
\end{figure*}

\begin{figure}[!htb]
\centering
\includegraphics[scale=0.7]{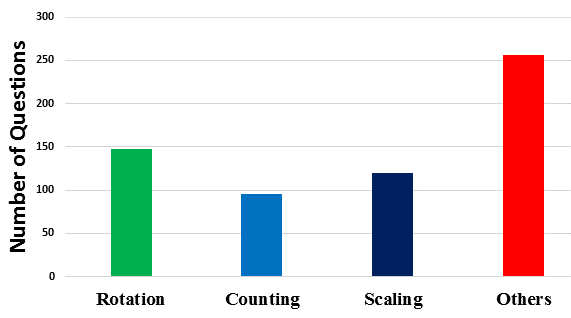}
\caption{Distribution of different DR problems in our dataset.}\label{fig:dataset3}
\end{figure}

We have collected images of diagrammatic reasoning from the web and prepared a dataset of $4 \times 1$ diagrammatic reasoning problems. The dataset contains 619 number of problems. We have categorized these problems into four groups, namely (i) Rotation (RT), (ii) Counting (CT), (iii) Shape Scaling (SS), and (iv) Other Type (OT). Figure~\ref{fig:dataset1} depicts one sample question with possible answers from each category and Figure~\ref{fig:dataset3} depicts the distribution of the problems across various categories in our dataset.

\section{KLSTM for Solving $4\times1$ DR Problems}
\begin{figure*}[!htb]
\begin{center}
\includegraphics[scale=0.4]{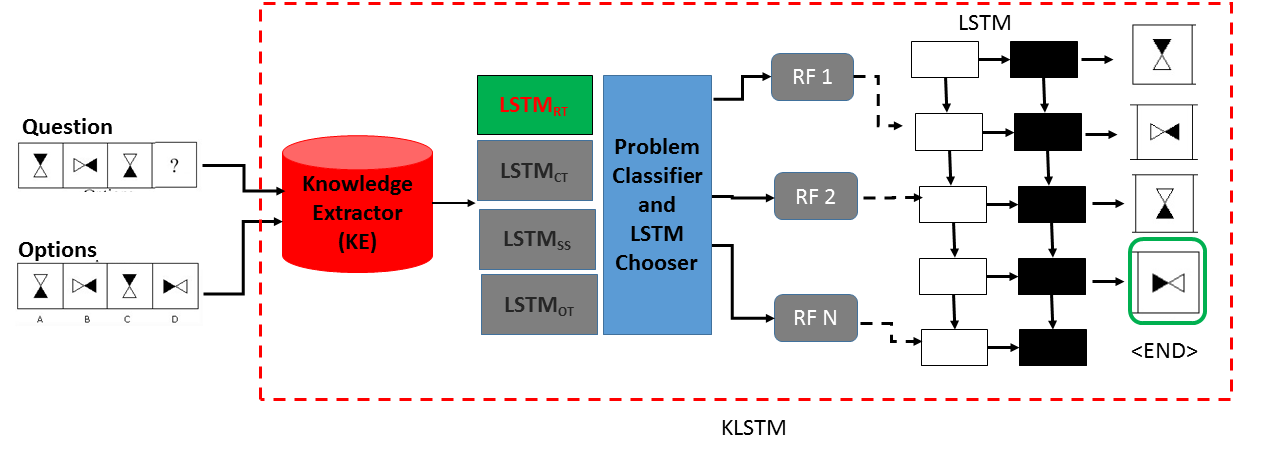}
\end{center}
\caption{Architecture of the proposed DR solving framework with various components. We take question sequence and the options as input and construct a knowledge base. Finally, it predicts the best possible option out of the four input options and produces a complete sequence of four patterns/images. The framework consist of a rule-base problem classifier and a set of LSTM similar to~\cite{vinyals2015show}. The input of the LSTM are relative features (RF).}
\label{fig:proposed}
\end{figure*}

The proposed DR solving method is based on a set of features and rules. We have introduced a supervised and ruled-based method to extract relational features (RF) of image sequences.
The proposed method consists of two major steps. During the first-level of processing, the question and options are passed through a knowledge acquisition tool to construct the knowledge base. The knowledge consists of a set of image features extracted from individual image and a set of relational features extracted from the sequence of images in the question. Next, the problem type is identified using a rule-based method. Finally, the features are passed through a Knowledge-specific Long Short Term Memory (KLSTM) to predict the possible output pattern or image. Figure~\ref{fig:proposed} depicts the proposed framework in details. The KLSTM network consists of (i) a Knowledge acquisition module, (ii) a set of LSTM, and (iii) a problem classifier and LSTM chooser module.

The problem space (P) is defined in~(\ref{eq:problem}), where the question contains a set of images $(Q)=\{I_1,I_2,I_3\}$ and the given options are grouped in another set $(O)=\{I_4,I_5,I_6,I_7\}$. Diagrammatic reasoning is to predict the answer image such that $I_{answer} \in O$. First, we represent the problem using a high-level knowledge structure. This is carried out as follows. Domain knowledge of human experts (rules) are used to understand the relation among a sequence of images. Next, a knowledge base $(K)$ is constructed. That expert opinions (rules) are integrated with the system to solve visual reasoning problems. The method is presented in Algorithm~\ref{alg:solve}.

\begin{equation}
\label{eq:problem}
P=\{I_1,I_2,I_3,...,I_7\}
\end{equation}

\begin{algorithm}
\caption{ Diagrammatic Reasoning } \label{alg:solve}
\begin{algorithmic}[1]
\Require {Problem Space as defined in~(\ref{eq:problem})}
\Ensure {$I_{answer}$, where $I_{answer}\in O$}
\State Extract knowledge base $(\kappa)$ from the training data
\State{$Q_{cat}$ =} Classify (Q, K), where $category=\{$RT, CT, SS, OT $\}$
\State $I_{answer}=LSTM_{j}(Q,O)$, where $j=category$
\State Return $I_{answer}$
\end{algorithmic}
\end{algorithm}

\textbf{Knowledge Acquisition:} Knowledge acquisition is carried out during training. The knowledge base $(\kappa)$ is extracted from a set of training samples, i.e. problems used in training. First, the shapes in each image in $P$ and number of similar shapes are extracted using YOLO~\cite{redmon2018yolov3}. YOLO is fast and the accuracy of the method is good enough in our context. We then introduce a new feature for solving digrammatic reasoning problems. The feature is referred to as the relational feature. Unlike image-based features such as color, texture, shape or edge that are typically used in various computer vision applications, we have extracted three relational features, namely rotation $(\rho)$, counts $(\chi)$, and scaling $(\sigma)$ from the set of the given images. The feature-set is given in ~(\ref{eq:knowledge}). Various components of the feature-set (k) are described hereafter.

\begin{equation}
\label{eq:knowledge}
\kappa=<\rho(I_k),\chi(I_k),\sigma(I_k)>,\forall k, k \in P
\end{equation}

\textbf{Shape Detection:} Each image in $P$ is passed through a deep learning module to extract the shapes. We have considered common geometrical shapes such as circle, triangle, rectangle, square, diamond, star, hexagon that are usually present in various DR problems. All the shapes are classified as either empty (only edges) or filled. YOLO~\cite{redmon2018yolov3} has been found to be a good binary classifier as compared to Resnet50/101~\cite{he2016deep}, VGG 16~\cite{simonyan2014very}, or GoogleNet~\cite{szegedy2015going}.

\begin{figure}[!htb]
\centering
\includegraphics[scale=0.6]{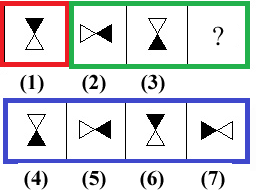}
\caption{We represent the rotation problem as a set of 7 images or patterns. In rotation problems, we consider the first image (red) as the reference image with $0\degree$ rotation and extract the rotation relation of other images.}
\label{fig:rotation}
\end{figure}

\begin{figure*}[!htb]
\centering
\includegraphics[scale=0.4]{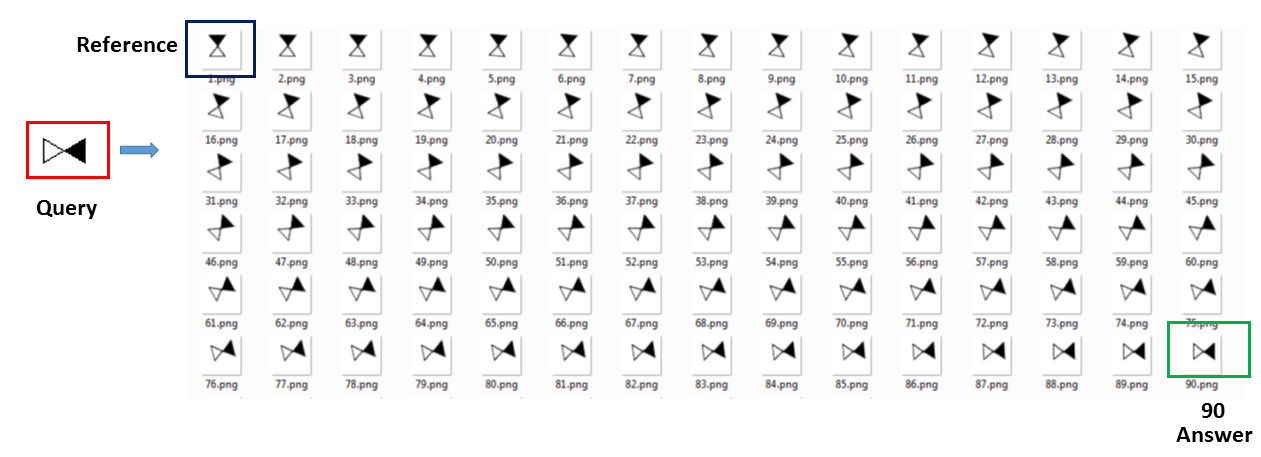}
\caption{Sample images after applying rotations on the first image of P. Depiction of how a possible match is found at $90\degree$ for a given query image.}
\label{fig:rotation_360}
\end{figure*}

\textbf{Rotation:} In a typical rotation diagrammatic reasoning problem (Figure~\ref{fig:rotation}), the solution lies in rotating the figure correctly to complete the sequence. We assume the first image ($I_1$) as the reference with a rotation of $0\degree$.  All other images $(I_2,...,I_7)$ are expressed using rotation angle with respect to the reference image.
To achieve this, 360 number of images are generated by incrementally rotating the base image by $1\degree$ . A few samples of the rotated images corresponding to the DR problem described in Figure~\ref{fig:rotation} are shown in Figure~\ref{fig:rotation_360}. This set is denoted by $R = \{I_1, I_2, ..., I_{360}\}$. The similarity score $(\psi)$ is defined in~(\ref{eq:rot1}). This score has been estimated between a query image and all images of $R$ using ResNet50, where $I_j$ is query image and $I_k$ is the image in $R$.
\begin{equation}
  \psi_{j} = Similarity(I_j,I_k)
  \label{eq:rot1}
\end{equation}

The relative rotation $\rho(I_k)$ of each image of $P$ is then extracted with respect to each image $I_j$ belonging to $R$. If the images in $P$ are different from each other, we categorize the question as non-rotation problem and a flag NA is set. A threshold has been used to decide about the success of matching. $\rho(I_k)$ is set to the value of rotation if the matching score returned by ResNet50 is above the threshold. However, in the event of multiple images being categorized above the threshold, the image that gives the highest value, is selected and its rotation angle is taken as the final input. In the event that none is found suitable, the problem is categorized as non-rotation digametic reasoning problem.

The relative rotations of the diagrammatic reasoning problem depicted in Figure~\ref{fig:dataset1}(a) are $\{0\degree, 90\degree, 180\degree \}$ for the options in the question and $\{180\degree, 270\degree, 0\degree, 90\degree \}$ for the options in the answer.

\begin{figure}[!htb]
\centering
\includegraphics[scale=0.4]{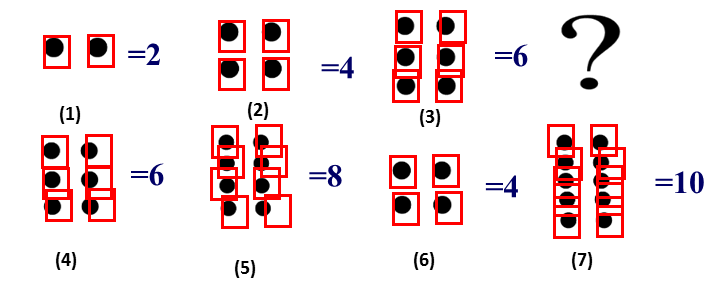}
\caption{A typical counting DR problem with 7 pictures. The first three patterns represent the sequence given in the question {2, 4, 6, ?} and the next four patterns represent the options for the probable answer with {8} as the correct option.}
\label{fig:count}
\end{figure}

\textbf{Counting:}
Counting is a reasoning problem where the solution is to extract the correct number of shapes present in the problem sequence. First, the shapes are detected and the number of same shapes is estimated. For example, Figure~\ref{fig:count} depicts a typical filled circle detection and counting using YOLO.
Each image of the problem space is expressed using the count of shapes in a sequence as $\{2,4,6,?\}$. The predicted missing number is from the set $\{6,8,4,10\}$.

\textbf{Scaling:} Relative scaling $(\sigma)$ is extracted from the bounding box of the detected shapes. First, the bounding boxes are extracted from the shapes in $P$. Next, similar sized shapes are grouped using unsupervised density-based spectral clustering with application to noise (DBSCAN)~\cite{tran2013revised}. The groups are then rearranged in order of labels such that $L_1<L_2...<L_n$. These groups can be labelled as large, medium, small and tiny for a typical $4\times1$ DR problem. The process of grouping and labelling of shapes is described in Algorithm~\ref{alg:scale}.

\begin{figure}[!htb]
\centering
\includegraphics[scale=0.5]{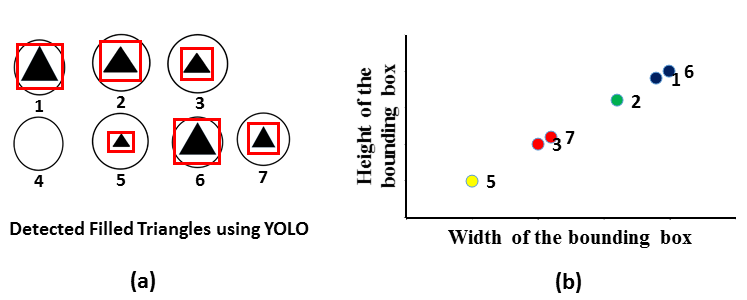}
\caption{(a) Detection of shapes archived by YOLO. (b) The bounding boxes are grouped using DBSCAN. Each color represents a group of same size shapes.}
\label{fig:scale}
\end{figure}

\begin{algorithm}
\caption{Diagrammatic Reasoning  for Scaling Problems}
\label{alg:scale}
\begin{algorithmic}[1]
\Require {Problem Space (P) as defined in equation~(\ref{eq:problem})}
\Ensure {Relational scaling $(\sigma)$ of each image}
\State S=DetectShapes($I_k$), $\forall k,I_k \in P$
\State ShapeGroup=DBSCAN(S)
\State Rearrange group and assign label $L_1,L_2,...,L_n$, where $L_1<L_2...<L_n$
\State $\sigma_k$= Shape Label$(I_k)$
\State Return $\sigma$
\end{algorithmic}
\end{algorithm}
Figure~\ref{fig:scale}(a) depicts a DR problem where size of the pattern is used as a clue for the solution. The DBSCAN algorithm has identified four classes or groups, where the problem has been expressed as $\{<Very Large>,<Large>,<Small>,?\}$, and the solution options are $\{<Nil>,<Tiny>,<Very Large>,<Small>\}$.

\textbf{Representation of Knowledge Base:} For a given problem space $P$, the shapes are detected and the relational features (RF) are extracted as mentioned earlier. The knowledge base consists of four sets, namely shapes, rotation $(\rho)$, counting $(\chi)$, and scaling $(\sigma)$. Shapes store information about the structures and other sets represent various components of the relational features. Table~\ref{tbl:kb} shows the knowledge extracted from four different $4 \times 1$ problems.
\begin{table*}[!htb]
\center
\caption{Typical examples of knowledge base extracted using the features described earlier}
\label{tbl:kb}
\resizebox{14cm}{!}{%
\begin{tabular}{@{}ll@{}}
\toprule
\textbf{DR Problem}                & \textbf{Constructed Knowledge Base} \\ \midrule
\begin{minipage}{.3\textwidth}
      \includegraphics[width=30mm, height=20mm]{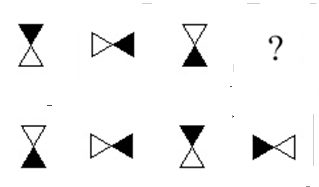}
    \end{minipage}

 & \begin{tabular}[c]{@{}l@{}}$Shapes=\{\textit{Filled triangle}, Triangle\}$\\$\rho=\{0\degree,90\degree,180\degree,180\degree,90\degree,0\degree,270\degree\}$\\ $\chi=\{<1,1>\,<1,1>\,<1,1>,<1,1>,<1,1>,<1,1>,<1,1>\}$\\ $\sigma=\{<N,N>,<N,N>,<N,N>,<N,N>,<N,N>,<N,N>,<N,N>\},\textit{where N is Normal.}$\end{tabular}              \\ \bottomrule
\begin{minipage}{.3\textwidth}
      \includegraphics[width=30mm, height=20mm]{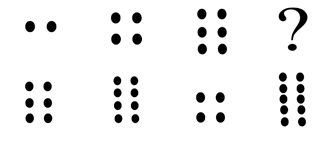}
    \end{minipage}

& \begin{tabular}[c]{@{}l@{}}$Shapes=\{\textit{Filled circle}\}$\\$\rho=\{NA\}$\\ $\chi=\{<2>,<4>,<6>,<6>,<8>,<4>,<10>\}$\\ $\sigma=\{<AN>,<AN>,<AN>,<AN>,<AN>,<AN>,<AN>\}, \textit{where AN is All Normal.}$\end{tabular}              \\ \midrule

\begin{minipage}{.3\textwidth}
      \includegraphics[width=30mm, height=20mm]{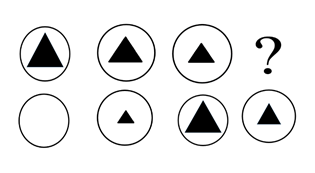}
    \end{minipage}

 & \begin{tabular}[c]{@{}l@{}}$Shapes=\{\textit{Filled triangle}\}$\\$\rho`=\{NA\}$\\ $\chi=\{<1>,<1>,<1>,<1>,<1>,<1>,<1>\}$
 \\ $\sigma=\{<Very Large>,<Large>,<Small>,<Nil>,<Tiny>,<Very Large>,<Small>\}$\end{tabular}              \\ \midrule

\begin{minipage}{.3\textwidth}
      \includegraphics[width=30mm, height=20mm]{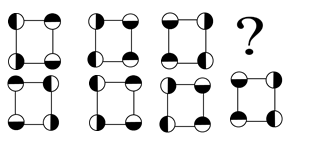}
    \end{minipage}

 & \begin{tabular}[c]{@{}l@{}}$Shapes=\{\textit{Filled triangle}\}$\\$\rho=\{NA\}$\\ $\chi=\{<4>,<4>,<4>,<4>,<4>,<4>,<4>\}$\\ $\sigma=\{<AN>,<AN>,<AN>,<AN>,<AN>,<AN>,<AN>\}, \textit{where AN is All Normal.}$\end{tabular}              \\ \bottomrule
\end{tabular}
}
\end{table*}

\begin{figure}[!htb]
\centering
\includegraphics[scale=0.45]{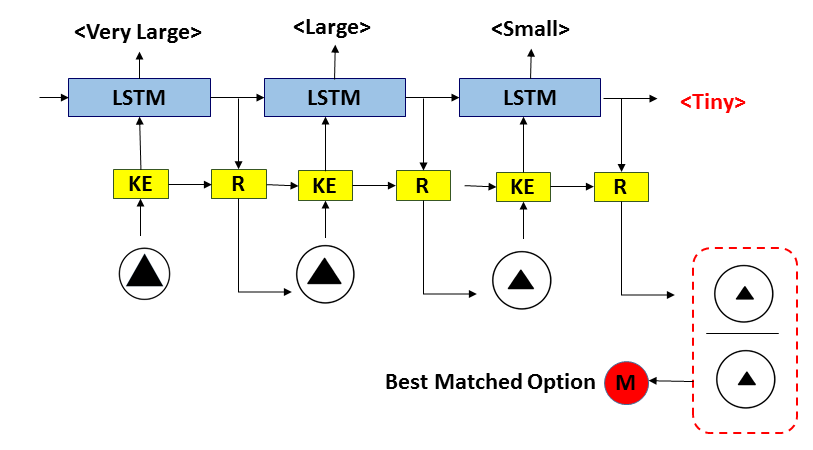}
\caption{Prediction model for solving Category 1 problems. KE: Knowledge Extractor, R: Representor, M: Matching Module.}
\label{fig:cat1}
\end{figure}

\textbf{Classification and Solving:} Final stage is to learn the pattern from the question images and predict the correct answer from the given options. At the beginning, the relational features (RF) are extracted from all training samples. Next, four independent LSTMs corresponding to  $\rho,\chi,\sigma$ and unknown problems are trained to build the prediction model. In the testing phase, a similar knowledge base is extracted from the test sample. Next, a rule-based method as described in Algorithm~\ref{alg:clasify} is applied to classify the problem as Category 1 (RT or CT or SS) or Category 2 (OT). In the case of Category 1, a variation of LSTM is used as proposed in~\cite{vinyals2015show}. The method has been used to generate a caption from the images. Rather than using the conventional image-based features~\cite{vukotic2017one,liu2017video}, we have used relational features (RF) extracted by the knowledge extractor. The method is depicted in Figure~\ref{fig:cat1}, where the knowledge extractor (KE) is the process of extracting RFs and representor (R) is the image with a set of relational feature.
\begin{algorithm}
\caption{ Classification of Diagrammatic Reasoning Problem}
\label{alg:clasify}
\begin{algorithmic}[1]
\Require {relational features of the problem}
\Ensure {Problem Class}
\If {$\rho \neq NA$ or $\rho$ are different}
\State $P=RT$
\ElsIf {$\chi \neq NA$ or $\chi$ are different}
\State $P=CT$
\ElsIf {$\sigma \neq NA$ or $\chi$ are different}
\State $P=SS$
\Else
\State $P=OT$
\EndIf
\State Return $P$
\end{algorithmic}
\end{algorithm}

\begin{figure}[!htb]
\centering
\includegraphics[scale=0.5]{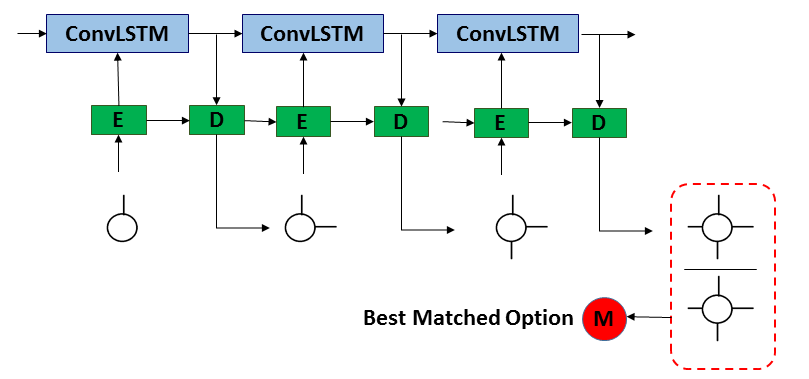}
\caption{Interpolation model for solving Category 2 problems. E: Encoder, D: Decoder, M: Matching Network.}
\label{fig:ot}
\end{figure}

Unknown problems (Category 2) are solved by the variation of LSTM called Flexible Spatio-Temporal Network (FSTN) proposed in~\cite{lu2017flexible}. Originally the method predicts the future video frames from a set of observed sequences. In this method, image-based features are sequentially passed through a LSTM in encoding/decoding manner. Figure~\ref{fig:ot} depicts the method in details. The method consists of encoders (E), decoders (D), and a matching network (M). The network is trained using several image sequences.
Table~\ref{tbl:predict} shows reference feature prediction of the problems shown in Table~\ref{tbl:kb}.

\begin{table*}[!htb]
\center
\caption{Details of the predicted knowledge and answers}
\label{tbl:predict}
\begin{tabular}{@{}llll@{}}
\toprule
\textbf{Predicted Answer}                & \textbf{Predicted Knowledge} & \textbf{Detected Category} & \textbf{Correct?}\\ \midrule
\begin{minipage}{.3\textwidth}
      \includegraphics[width=30mm, height=20mm]{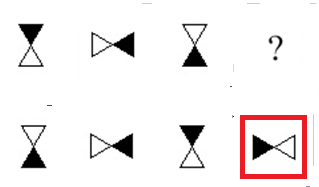}
    \end{minipage}

 & \begin{tabular}[c]{@{}l@{}}$Shapes=\{\textit{Filled triangle}, Triangle\}$\\$\rho=\{270\degree\}$\\ $\chi=\{<1,1>\}$\\ $\sigma=\{<N,N>\}$\end{tabular}              & Category 1 (RT)& Yes\\ \bottomrule
\begin{minipage}{.3\textwidth}
      \includegraphics[width=30mm, height=20mm]{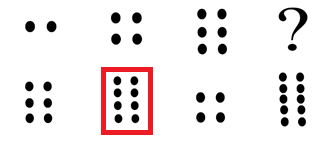}
    \end{minipage}

& \begin{tabular}[c]{@{}l@{}}$Shapes=\{\textit{Filled circle}\}$\\$\rho=\{NA\}$\\ $\chi=\{<8>\}$\\ $\sigma=\{<AN>\}$
\end{tabular}              & Category 1 (CT)& Yes\\ \midrule

\begin{minipage}{.3\textwidth}
      \includegraphics[width=30mm, height=20mm]{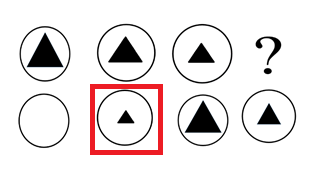}
    \end{minipage}

 & \begin{tabular}[c]{@{}l@{}}$Shapes=\{\textit{Filled triangle}\}$\\$\rho=\{NA\}$\\ $\chi=\{<1>\}$
 \\ $\sigma=\{<Tiny>\}$
 \end{tabular}              & Category 1 (SS)& Yes\\ \midrule

\begin{minipage}{.3\textwidth}
      \includegraphics[width=30mm, height=20mm]{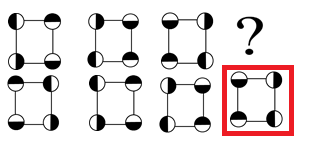}
    \end{minipage}

& NA & Category 2 (OT)& No\\ \bottomrule
\end{tabular}
\end{table*}

\section{Experiments using Baselines}
We present the experiment results in this section. The first step of the method is to detect shapes from a given image. We have experimented with state-of-the-art convolutional networks including ResNet50, ResNet101~\cite{he2016deep}, VGG16~\cite{simonyan2014very}, GoogleNet~\cite{szegedy2015going} and YOLO~\cite{redmon2018yolov3}. YOLO has been found to be the best architecture for the present case. 70\% of the data have been used for training and 30\% for testing across all experiments. We have performed 10 folds cross validation and reported the average results. Table~\ref{tbl:shapedetect} summarizes the shape detection results.

\begin{figure}[!htb]
\centering
\includegraphics[scale=0.4]{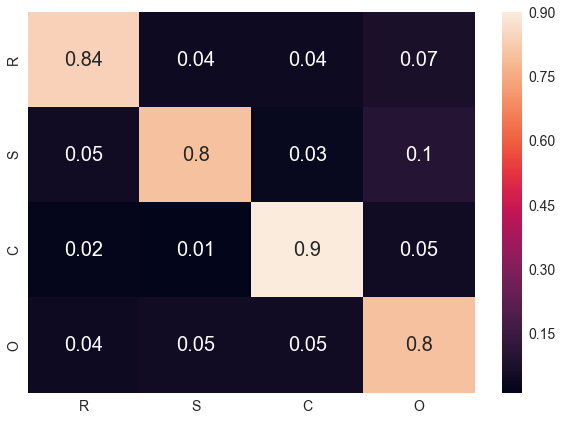}
\caption{\textbf{The confusion matrix for classifying DR problem.} R: RT, C: CT, S: SS, O: OT.}
\label{fig:classify}
\end{figure}

\begin{table}[!htb]
\caption{Results of shape detection}
\label{tbl:shapedetect}
\center
\begin{tabular}{@{}ll@{}}
\toprule
Algorithm                       & \textbf{Accuracy} \\ \midrule
\textbf{ResNet50 (baseline)}               & 57.19             \\
\textbf{ResNet101~\cite{he2016deep}}            & 62.19             \\
\textbf{VGG16~\cite{simonyan2014very}}              & 71.11             \\
\textbf{GoogleNet~\cite{szegedy2015going}} & 77.22             \\
\textbf{YOLO~\cite{redmon2018yolov3}}                   & 86.76             \\ \bottomrule
\end{tabular}
\end{table}
In the next stage, classification of the problem type has been carried out. The confusion matrix for four types of problems is depicted in Figure~\ref{fig:classify}. It is observed that the proposed method can successfully classify the problems with reasonably high accuracy.

\begin{figure*}[!htb]
\centering
\includegraphics[scale=0.5]{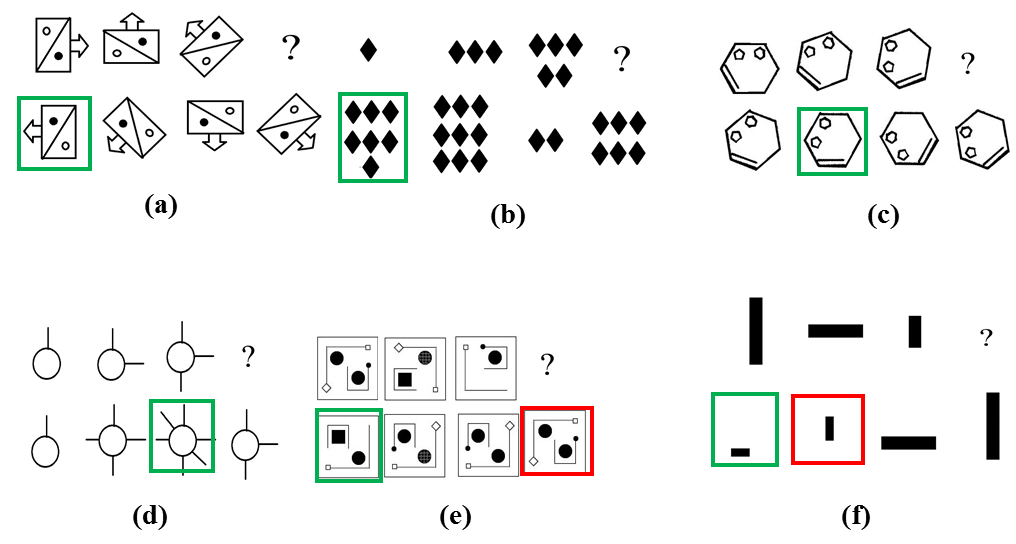}
\caption{A few samples of the proposed DR dataset. Green boxes represent ground truths and correctly solved problems, red boxes represent wrongly predicted answers.}
\label{fig:res}
\end{figure*}

We have carried out several experiments to understand the behavior of the DR problem solver. We have taken image-based features as baseline and applied state-of-the-art recurrent neural network (RNN)  to solve the reasoning problems. The results are summarized in Table~\ref{tbl:solveaccuracy}. Figure~\ref{fig:res} depicts some success and failure cases.

\begin{table*}[!htb]
\center
\caption{Comparative results of DR problem solving}
\label{tbl:solveaccuracy}
\resizebox{14cm}{!}{%
\begin{tabular}{@{}llllll@{}}
\toprule
\textbf{Algorithm}          & \textbf{Rotation (RT)} & \textbf{Counting (CT)} & \textbf{Scaling (SS)} & \textbf{Other (OT)} & \textbf{Average} \\ \midrule
\textbf{Image+LSTM (baseline)}  & 57.12                  & 42.13                  & 62.11                 & 36.5                & 49.46            \\
\textbf{Image+Encoder/Decoder~\cite{vukotic2017one}} & 62.11                  & 41.12                  & 61.11                 & 37.89               & 50.55            \\
\textbf{Image+Deep feature~\cite{lan2014hierarchical}}        & 64.39                  & 47.19                  & 41.91                 & 42.86               & 49.08            \\
\textbf{Image+RNN~\cite{bengio2015scheduled}}        & 56.80                  & 41.19                  & 54.91                 & 32.20               & 46.27            \\
\textbf{Image+FSTN~\cite{lu2017flexible}}        & 66.11                  & 37.19                  & 66.91                 & 34.90               & 51.27            \\
\textbf{RF+FSTN~\cite{lu2017flexible}}        & 51.16                  & 52.19                  & 46.34                 & 42.86               & 48.13            \\
\textbf{Proposed KLSTM}           & 75.87                  & 76.22                  & 73.41                 & 66.91               & 73.10            \\ \bottomrule
\end{tabular}
}
\end{table*}
\section{Conclusion}
In this paper, we have introduced a new dataset for solving diagrammatic reasoning (DR) problems using machine learning and computer vision. The dataset can open up new challenges to the vision community. We have experimented with several state-of-the-art learning frameworks to solve typical $4\times1$ DR problem. It has been observed that the image-based analysis usually fails to answer correctly in many cases. We have introduced a new feature set called relational feature. A rule-based learning with the help of LSTM has been used to classify the DR questions. Results reflect that the proposed rule-based method outperforms existing image-based analysis.

It has been observed that simple rules defined in this work may not be sufficient to solve all types of DR problems. Complicated rules need to be defined and we may need to redefine the feature-set for solving complex DR problems. Mainly, other types (OT) DR problems need further attention of the research community.
\section*{Acknowledgement}
\textbf{Funding:} We gratefully acknowledge the support of NVIDIA Corporation with the donation of the Quadro P5000 GPU used for this research.
\\
\textbf{Conflict of interest:} The authors declare that there is
no conflict of interest regarding the publication of this paper.\\
\textbf{Ethical approval:} This article does not contain any
studies with human participants or animals performed by any of the
authors. Informed consent: Informed consent was obtained from all
individual participants included in the study.
\section*{References}
\bibliography{egbib}

\begin{thebibliography}{}
\expandafter\ifx\csname url\endcsname\relax
  \def\url#1{\texttt{#1}}\fi
\expandafter\ifx\csname urlprefix\endcsname\relax\def\urlprefix{URL }\fi
\expandafter\ifx\csname href\endcsname\relax
  \def\href#1#2{#2} \def\path#1{#1}\fi

\end{thebibliography}


\begin{thebibliography}{10}
\expandafter\ifx\csname url\endcsname\relax
  \def\url#1{\texttt{#1}}\fi
\expandafter\ifx\csname urlprefix\endcsname\relax\def\urlprefix{URL }\fi
\expandafter\ifx\csname href\endcsname\relax
  \def\href#1#2{#2} \def\path#1{#1}\fi

\bibitem{kushman2014learning}
N.~Kushman, Y.~Artzi, L.~Zettlemoyer, R.~Barzilay, Learning to automatically
  solve algebra word problems, in: Proceedings of the 52nd Annual Meeting of
  the Association for Computational Linguistics (Volume 1: Long Papers),
  Vol.~1, 2014, pp. 271--281.

\bibitem{antol2015vqa}
S.~Antol, A.~Agrawal, J.~Lu, M.~Mitchell, D.~Batra, C.~Lawrence~Zitnick,
  D.~Parikh, Vqa: Visual question answering, in: Proceedings of the IEEE
  international conference on computer vision, 2015, pp. 2425--2433.

\bibitem{yang2016stacked}
Z.~Yang, X.~He, J.~Gao, L.~Deng, A.~Smola, Stacked attention networks for image
  question answering, in: Proceedings of the IEEE conference on computer vision
  and pattern recognition, 2016, pp. 21--29.

\bibitem{noh2016image}
H.~Noh, P.~Hongsuck~Seo, B.~Han, Image question answering using convolutional
  neural network with dynamic parameter prediction, in: Proceedings of the IEEE
  conference on computer vision and pattern recognition, 2016, pp. 30--38.

\bibitem{johnson2017inferring}
J.~Johnson, B.~Hariharan, L.~van~der Maaten, J.~Hoffman, L.~Fei-Fei, C.~L.
  Zitnick, R.~B. Girshick, Inferring and executing programs for visual
  reasoning., in: ICCV, 2017, pp. 3008--3017.

\bibitem{hu2017learning}
R.~Hu, J.~Andreas, M.~Rohrbach, T.~Darrell, K.~Saenko, Learning to reason:
  End-to-end module networks for visual question answering, CoRR,
  abs/1704.05526 3.

\bibitem{hu2017modeling}
R.~Hu, M.~Rohrbach, J.~Andreas, T.~Darrell, K.~Saenko, Modeling relationships
  in referential expressions with compositional modular networks, in: Computer
  Vision and Pattern Recognition (CVPR), 2017 IEEE Conference on, IEEE, 2017,
  pp. 4418--4427.

\bibitem{johnson2017clevr}
J.~Johnson, B.~Hariharan, L.~van~der Maaten, L.~Fei-Fei, C.~L. Zitnick,
  R.~Girshick, Clevr: A diagnostic dataset for compositional language and
  elementary visual reasoning, in: Computer Vision and Pattern Recognition
  (CVPR), 2017 IEEE Conference on, IEEE, 2017, pp. 1988--1997.

\bibitem{balanced_vqa_v2}
Y.~Goyal, T.~Khot, D.~Summers{-}Stay, D.~Batra, D.~Parikh, Making the {V} in
  {VQA} matter: Elevating the role of image understanding in {V}isual
  {Q}uestion {A}nswering, in: Conference on Computer Vision and Pattern
  Recognition (CVPR), 2017.

\bibitem{gavornik2014learned}
J.~P. Gavornik, M.~F. Bear, Learned spatiotemporal sequence recognition and
  prediction in primary visual cortex, Nature neuroscience 17~(5) (2014) 732.

\bibitem{lu2017flexible}
C.~Lu, M.~Hirsch, B.~Sch{\"o}lkopf, Flexible spatio-temporal networks for video
  prediction, in: Proceedings of the IEEE Conference on Computer Vision and
  Pattern Recognition, 2017, pp. 6523--6531.

\bibitem{liu2017video}
Z.~Liu, R.~A. Yeh, X.~Tang, Y.~Liu, A.~Agarwala, Video frame synthesis using
  deep voxel flow., in: ICCV, 2017, pp. 4473--4481.

\bibitem{vukotic2017one}
V.~Vukoti{\'c}, S.-L. Pintea, C.~Raymond, G.~Gravier, J.~C. van Gemert,
  One-step time-dependent future video frame prediction with a convolutional
  encoder-decoder neural network, in: International Conference on Image
  Analysis and Processing, Springer, 2017, pp. 140--151.

\bibitem{vinyals2015show}
O.~Vinyals, A.~Toshev, S.~Bengio, D.~Erhan, Show and tell: A neural image
  caption generator, in: Proceedings of the IEEE conference on computer vision
  and pattern recognition, 2015, pp. 3156--3164.

\bibitem{redmon2018yolov3}
J.~Redmon, A.~Farhadi, Yolov3: An incremental improvement, arXiv preprint
  arXiv:1804.02767.

\bibitem{he2016deep}
K.~He, X.~Zhang, S.~Ren, J.~Sun, Deep residual learning for image recognition,
  in: Proceedings of the IEEE conference on computer vision and pattern
  recognition, 2016, pp. 770--778.

\bibitem{simonyan2014very}
K.~Simonyan, A.~Zisserman, Very deep convolutional networks for large-scale
  image recognition (2015) 19--36.

\bibitem{szegedy2015going}
C.~Szegedy, W.~Liu, Y.~Jia, P.~Sermanet, S.~Reed, D.~Anguelov, D.~Erhan,
  V.~Vanhoucke, A.~Rabinovich, Going deeper with convolutions, in: Proceedings
  of the IEEE conference on computer vision and pattern recognition, 2015, pp.
  1--9.

\bibitem{tran2013revised}
T.~N. Tran, K.~Drab, M.~Daszykowski, Revised dbscan algorithm to cluster data
  with dense adjacent clusters, Chemometrics and Intelligent Laboratory Systems
  120 (2013) 92--96.

\bibitem{lan2014hierarchical}
T.~Lan, T.-C. Chen, S.~Savarese, A hierarchical representation for future
  action prediction, in: European Conference on Computer Vision, Springer,
  2014, pp. 689--704.

\bibitem{bengio2015scheduled}
S.~Bengio, O.~Vinyals, N.~Jaitly, N.~Shazeer, Scheduled sampling for sequence
  prediction with recurrent neural networks, in: Advances in Neural Information
  Processing Systems, 2015, pp. 1171--1179.

\end{thebibliography}

\end{document}